# FastDTW is approximate and Generally Slower than the Algorithm it Approximates

Renjie Wu and Eamonn J. Keogh

**Abstract**—Many time series data mining problems can be solved with repeated use of distance measure. Examples of such tasks include similarity search, clustering, classification, anomaly detection and segmentation. For over two decades it has been known that the Dynamic Time Warping (DTW) distance measure is the best measure to use for most tasks, in most domains. Because the classic DTW algorithm has quadratic time complexity, many ideas have been introduced to reduce its amortized time, or to quickly approximate it. One of the most cited approximate approaches is FastDTW. The FastDTW algorithm has well over a thousand citations and has been explicitly used in several hundred research efforts. In this work, we make a surprising claim. In any realistic data mining application, the *approximate* FastDTW is much slower than the *exact* DTW. This fact clearly has implications for the community that uses this algorithm: allowing it to address much larger datasets, get exact results, and do so in less time.

———————————— ◆ ————————————

## 1 INTRODUCTION

Many time series data mining problems can be solved with repeated use of an appropriate distance measure. Examples of such tasks include similarity search, clustering, classification, anomaly detection, rule discovery, summarization and segmentation. It has long been believed that the Dynamic Time Warping (DTW) distance measure is the best measure to use in many domains, and recent extensive empirical "bake-offs" have confirmed this [1], [2], [3], [4], [5]. Because the DTW algorithm has time complexity that is quadratic in the length of the sequences, many ideas have been introduced to reduce its amortized time [3], or to quickly approximate it [1]. One of the most cited approximate approaches is FastDTW [6]. FastDTW works by creating an approximation of classic (full) DTW by computing DTW on a downsampled version of the data, then iteratively projecting the solution discovered onto an upsampled version and refining it.

At least *dozens*, but perhaps as many as *hundreds* of research efforts explicitly adopt FastDTW in order to gain scalability. The quotes below come from some representative works:

- "*In order to expedite the algorithm, we adopted fastDTW in our work*" [7].
- "(to) *minimize the computational complexity, we use a method called FastDTW*" [8].
- "*To increase the speed of the process, we employed a faster version of DTW, called FastDTW*" [9].
- "*FastDTW provides an efficient approximation to DTW*" [10].
- "*We used FastDTW to analyze the recorded accelerometer data for a first implementation of the gesture recognition*" [11].
- "*we use the FastDTW algorithm to automatically identify matching segments*" [12].
- "*the distances between the time series are computed using FastDTW*" [13].

To gauge how commonly used this algorithm is, consider the fact that at least five papers use the term FastDTW in their *title* [14], [15], [16], [17], [18].

In this work, we make a surprising claim. In any realistic setting, FastDTW is actually *slower* than DTW. Every paper that we are aware of that uses FastDTW would have obtained faster results by using simple DTW. Moreover, these results would have been exact (by definition), not approximate.

Clearly there are many other papers that made claims that did not pan out with the passage of time. Indeed, with the conceit of hindsight, the second author of this current work could not claim to be innocent of such transgressions. However, the FastDTW paper is very unusual in that the proposed algorithm is commonly used, especially by people outside the data mining community. That is to say, practitioners in medicine, bioengineering, industry, etc.

Moreover, there are many research efforts whose main goal is to improve FastDTW. For example, a recent paper summarizes its research contribution with: "*In the opinion of the authors, the disparity between the alignment inefficiency of the FastDTW algorithm and that of* (our algorithm) *is especially significant*." [19]. This effort also uses FastDTW as benchmark for accuracy: "*While* (our algorithm) *is slightly less accurate than FastDTW…*" [19]. However, the speed-up reported over FastDTW$_5$ is a factor of about five for time series of length 150. But as we will show, using off-the-shelf DTW would have dwarfed this apparent speedup.

The rest of this paper is organized as follows. In Section 2, we briefly review the necessary background mate-

- *R. Wu is with the Department of Computer Science and Engineering, University of California Riverside, Riverside, CA 92521. E-mail: rwu034@ucr.edu.*
- *E.J. Keogh is with the Department of Computer Science and Engineering, University of California Riverside, Riverside, CA 92521. E-mail: eamonn@cs.ucr.edu.*

rial and notation. Section 3 divides the similarity measurement task into four possibilities, which are empirically investigated. In Section 4, we show that even if FastDTW were faster than classic DTW, it is not clear to most people when it could fail to give a high-quality approximation. Section 5 summarizes our claims before we offer conclusions in Section 6.

## 2 Background and Notation

This review of DTW will be succinct, we encourage the interested reader to consult [2], [3] and the references therein for more information.

DTW reports the distance of two time series after optimally aligning them. DTW is computed by finding the minimum cost path in distance matrix $D$ of two time series $X$ and $Y$, where $D(i,j) = (X[i] - Y[j])^2 + \min\{D(i\text{-}1,j\text{-}1), D(i\text{-}1,j), D(i,j\text{-}1)\}$.

Since at least the 1970s, many practitioners have added *constraints* to the allowable warping paths. These constraints are normally denoted as $w$, which limits the number of cells (warping window) explored by DTW and the warping path allowed to deviate at most $w$ cells from the diagonal in computing $D$. Note that most papers report the warping constraint as a percentage of the length of time series, a practice we follow here. In this work we denote DTW with the constraint of $w$ as $cDTW_w$. Two special cases are worth noting. The case $cDTW_0$ is equivalent to the Euclidean distance, and the case $cDTW_{100}$ is equivalent to "unconstrained" or "Full" DTW.

It is important to correct a common misunderstanding here, even though a widely cited paper corrected it sixteen years ago [2]. Many people still believe that the purpose of using cDTW is to speed up the computation of DTW. However, this is only a happy side effect of using constraints on the warping path. The real purpose of using cDTW is that it is almost always more accurate because it prevents pathological warpings (see [2], [4]). An example of a pathological warping is when, say, a single heartbeat maps onto a dozen heartbeats. The use of cDTW with a suitable value of $w$, allows a short heartbeat to align to a longer heartbeat, but prevents this meaningless one-to-a-dozen alignment.

FastDTW is an approximation to Full DTW. A full exposition of FastDTW can be found in [6]. Briefly, FastDTW performs three steps recursively with a parameter radius ($r$). At each level of recursion, FastDTW downsamples the two time series being compared to half their length. Then FastDTW invokes itself to find the warping path of the two smaller (lower resolution) time series. Finally, a limited DTW is computed at the higher resolution. The warping window is the *neighborhood* of the projected warping path from lower resolution. The size of the neighborhood (i.e. the number of cells away from the projected warping path) is determined by $r$. Since FastDTW approximates Full DTW, $r$ can be seen as the tradeoff between precision and time: to achieve better accuracy of approximation, a larger $r$ is required; to reduce the running time of the algorithm, a smaller $r$ is necessary.

It is important to restate that $w$ and $r$ are not the same thing. The former is the parameter to give different maximum warping constraints, the latter is a parameter to control the tradeoff between accuracy and speed for FastDTW.

To be clear, we use $FastDTW_r$ for FastDTW with the radius of $r$ and $cDTW_w$ to denote cDTW with the warping window width of $w$.

In this work, we use $N$ to refer to the length of the time series being compared, $r$ for the radius of FastDTW, and $w$ for the user-specified warping constraint, given as a percentage of $N$. We use $W$ to refer to the *natural* amount of warping needed to align two random examples in a domain, also given as a percentage of $N$. This value can be difficult to know exactly, however there are often strong domain hints. For example, when aligning classical music performances, it is clear that there can be differences in timing between performances, however Kwon *et al.* [20] estimated that this is not more than 0.2 seconds. Thus, for a two-minute music performance this would suggest $W$ = 0.16%.

## 3 Four Cases in Similarity Measurement

With our notation established, in Table 1 we can consider the following exhaustive and exclusive matrix of possible settings in which DTW can be used.

TABLE 1
Four settings in which DTW can be used

| $N$ gets larger → | | |
|---|---|---|
| | **Case B**<br>Music performance, classical dance performance, seismic data | **Case D**<br><no obvious applications> |
| | **Case A**<br>Heartbeats, gestures, signatures, golf swings, gene expressions, gait cycles, star-light-curves, sign language words or phrases, bird song | **Case C**<br>Residential electrical power demand |
| | $W$ gets larger → | |

The boundaries between these four cases are somewhat subjective. For our purposes we will say that $N$ transitions from short to long somewhere around 1,000, and that $W$ transitions from narrow to wide somewhere around $W$ = 20%.

We can now consider the utility of FastDTW for each of these cases.

### 3.1 Case A: Short *N* and Narrow *W*

For this case, cDTW is unambiguously faster. Moreover, the original authors echo this point, writing in 2020 that "*If* (*W*) *is known beforehand to be* (small), *I recommend cDTW and do not recommend fastDTW*." [21].

To show how slow FastDTW can be compared to a vanilla iterative implementation of cDTW, consider Fig. 1. Here we consider the UWaveGestureLibraryAll dataset, which has exemplars of length 945, towards the long end of Case A. We consider all values of $w$ from 0 to 20%.

In this case, we know the best value of $W$ for this dataset, at least in the context of classification. The UCR archive notes that the error rate of $cDTW_0$ (i.e. Euclidean distance) is 0.052, that cDTW$_4$ minimizes the error to 0.034, and that cDTW$_{100}$ (i.e. Full DTW or unconstrained DTW) has a much higher error rate of 0.108.

It is worth discussing those results. The classification error rate of Full DTW is much higher than the error rate of constrained DTW. This continues to surprise people, but it has been known since at least 2004 [2]. It is sometimes referred to as the Ratanamahatana's observation "*a little warping is a good thing, but too much warping* (can be) *a bad thing*." [2].

It is important to recall that the two algorithms being compared are both implemented in the same language, running on the same hardware, performing the same task.

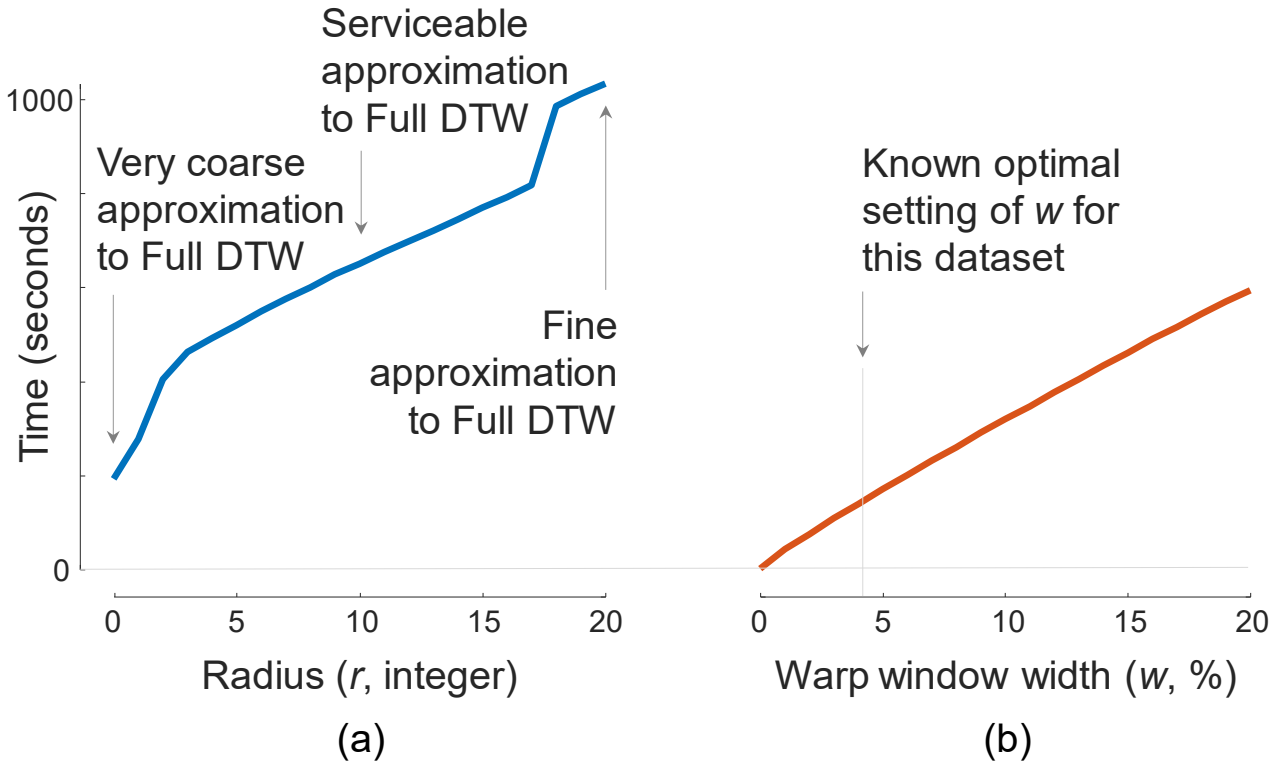

Fig. 1. A comparison of the time needed to compute all pairwise distances of the 896 training examples in UWaveGestureLibraryAll, (which requires (896×895)÷2 = 400,960 comparisons) for *r* = 0 to 20 for FastDTW (a) and for *w* = 0 to 20% for cDTW (b).

It is also important to state that we did not use any optimizations for cDTW. It is well known that when doing repeated measurements of DTW, say to find an object's nearest neighbor or to do nearest neighbor classification, one can avail of both lower bounding and early abandoning [2], [22]. Moreover, these ideas have be carefully optimized by the community for DTW. Using these ideas would have shaved at least two further orders of magnitude off the time for cDTW.

Note that our Fig. 1 (a) annotations of how well FastDTW approximates Full DTW are taken from the original paper. We do not make any comment on the quality of approximation here, other than to say that we assume the original claims are true. Thus Fig. 1 shows that for the optimal setting of *w* for this dataset, cDTW$_4$ is faster than the coarsest and fastest version of FastDTW. Moreover, even if we insisted on setting a larger value of *w*, up to 20, we can still exactly compute cDTW$_{20}$ as fast as we can compute a serviceable approximation to Full DTW, by using FastDTW$_{10}$. Thus, this experiment provides forceful evidence that at least for Case A, FastDTW is slower than using cDTW.

We believe that at least 99% of all uses of DTW in the literature fall into this case. One way to see this is to consider the distribution of *N* and *W* for the 128 datasets in the UCR Archive. This archive is clearly not a perfect representation of all datasets, all domains, and all problems. However, it is the largest such collection of labeled time series data in the world, and the optimal setting of *w* (which is our proxy for *W*) that maximizes classification accuracy, was computed by brute-force search [5]. Fig. 2 summarizes the data.

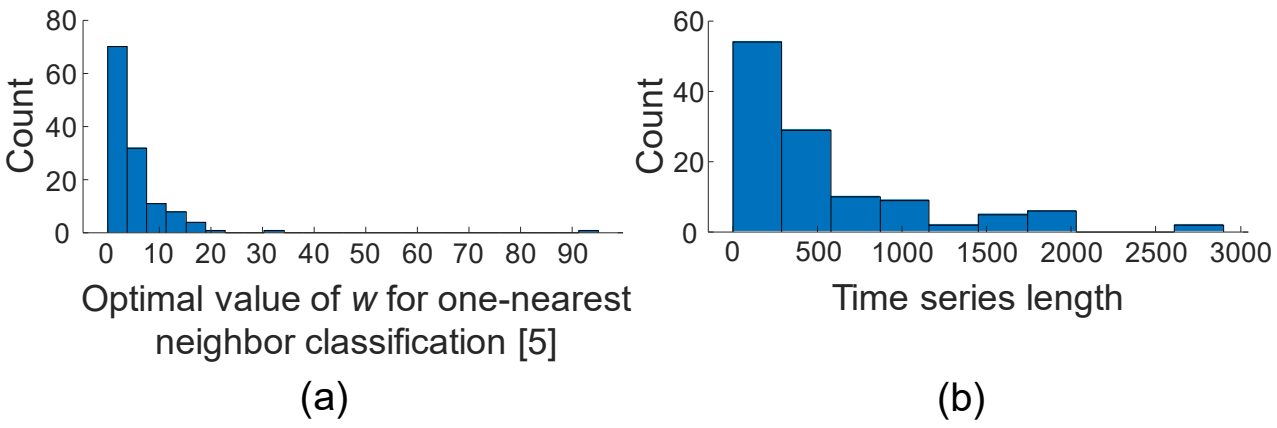

Fig. 2. (a) The distribution of optimal values for *w*, for the task of one-nearest neighbor classification, for 128 datasets. (b) The distribution of the lengths of these datasets.

These histograms show that majority of time series subsequences considered are less than 1,000 datapoints, and more importantly, the best value for *w* is rarely above 10%. Almost all uses of FastDTW fall into Case A [7], [8], [9], [10], [11], [12], [13], [14], [15], [16], [17], [18], and in every case the researchers using FastDTW would have been better off using classic cDTW, which would have been much faster, and *exact*.

Thus, the vast majority of readers of this work, who are reading this paper to decide if they should use cDTW or FastDTW, can stop reading here. Both the current authors, and the original authors of FastDTW [21], are recommending that you use cDTW. There is no disagreement or ambiguity for this case.

### 3.2 Case B: Long *N* and Narrow *W*

Case B considers the possibility of long time series with a low value for *W*. We have already hinted at one such possibility, musical performances, where the task is sometimes called score following or score alignment. Is cDTW faster here?

Let us perform an experiment. Imagine we align the exactly four-minute-long song "*Let It Be*" with a live version. For classical music, various papers have suggested values such as *W* = 0.16% [20]. Let us be much more liberal and assume that the live version can be up to two seconds ahead or behind at some point[1]. Thus, we set *w* = 0.83%. Music processing typically uses Chroma Features, which are normally sampled at 100Hz, thus we have a times series of length 24,000. To obtain a robust estimate we measured the time required for each algorithm one thousand times reporting the average. We find that:

- cDTW$_{0.83}$ takes 45.6 milliseconds.
- FastDTW$_{10}$ takes 238.2 milliseconds.
- FastDTW$_{40}$ takes 350.9 milliseconds.

Thus, for Case B we find no evidence of the utility of FastDTW.

### 3.3 Case C: Short *N* and Wide *W*

Case C considers the case where the *N* is short (say <1,000 data points), but *W* is large. There are no examples in the UCR archive, and a search of the literature does not suggest examples. However, the second author has a large collection of datasets, and after a significant effort man-aged to create a somewhat contrived situation/dataset.

[1] With apologies to Sir Paul McCartney, who has superb timing.

Imagine that a researcher decides to compare the first hour of electrical power demand each day in a residence (i.e. from midnight to 1am). Most of the time these would not be very similar under any measure. However, as shown in Fig. 3, we occasionally encounter a pattern that is similar, but only under the assumptions of Case C, where *N* is reasonably short (here 450 datapoints) but *W* is a large fraction of this value. Note that if we just use one pattern as a query on a sliding window of the entire year-long trace, the value of *W* would dramatically decrease, and we would be back in Case A.

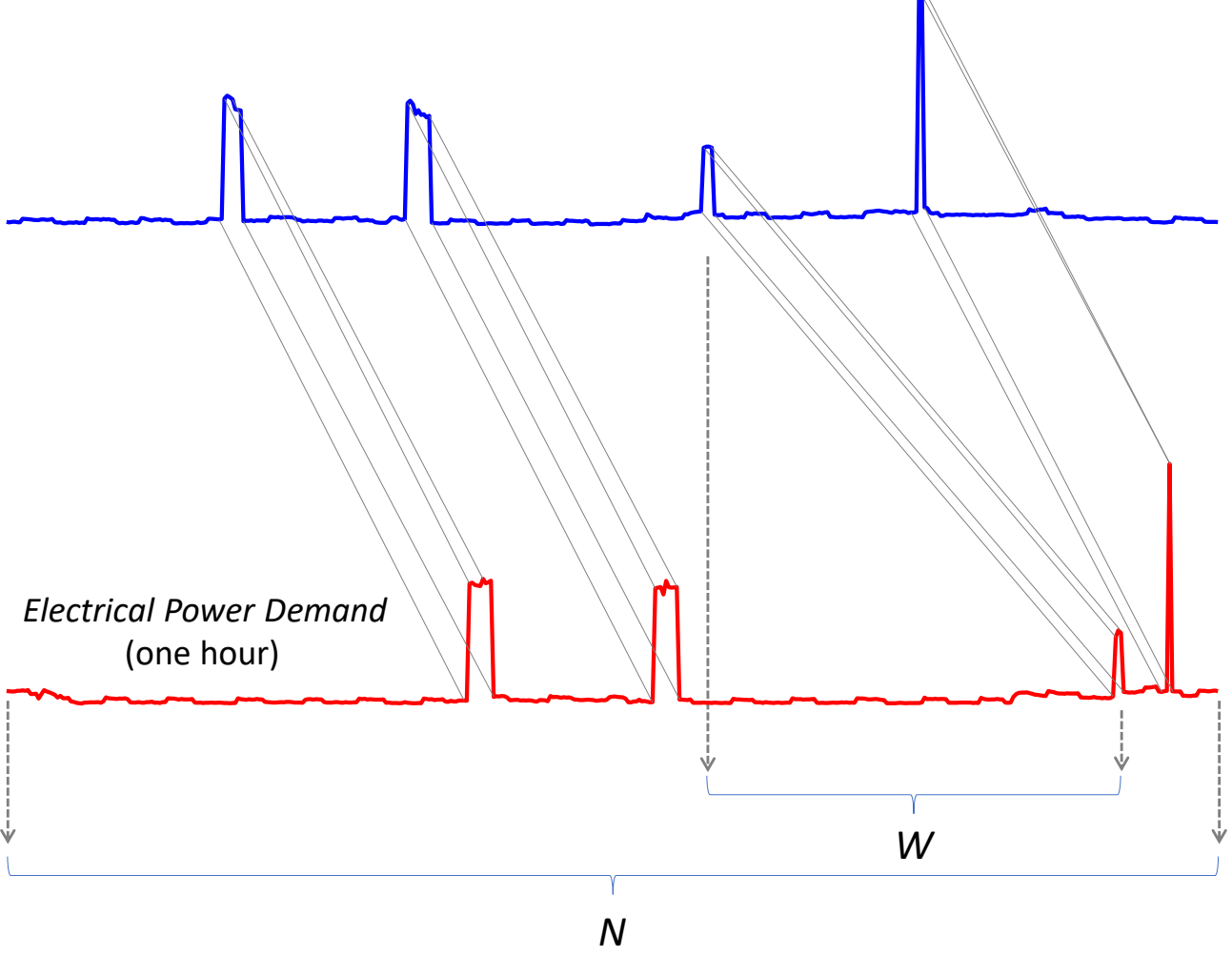

Fig. 3. An example of a case where *W* is a large faction of *N*. Two examples of the electrical power demand from midnight to 1:00 AM, sampled once per eight seconds. This conserved pattern reflects the program of a dishwasher. The owner may have programmed it to run after midnight, when the electrical power costs are cheaper in the UK.

Let us use the electrical power demand to motivate Case C. The natural value of *W* here is estimated by looking at the *maximum* difference in timing between corresponding pairs of peaks. This happens for the third pair, which differ by 153 datapoints. Given that time series are of length 450, that gives us an estimate of *W* = 34%, which to be conservative, we will round up to 40%. Thus, in Fig. 4, we repeat the type of experiment shown in Fig. 1, but consider time series of length 450, and *w* from 0 to 40%.

Since the timing for both algorithms does not depend on the data itself, we use random walk datasets.

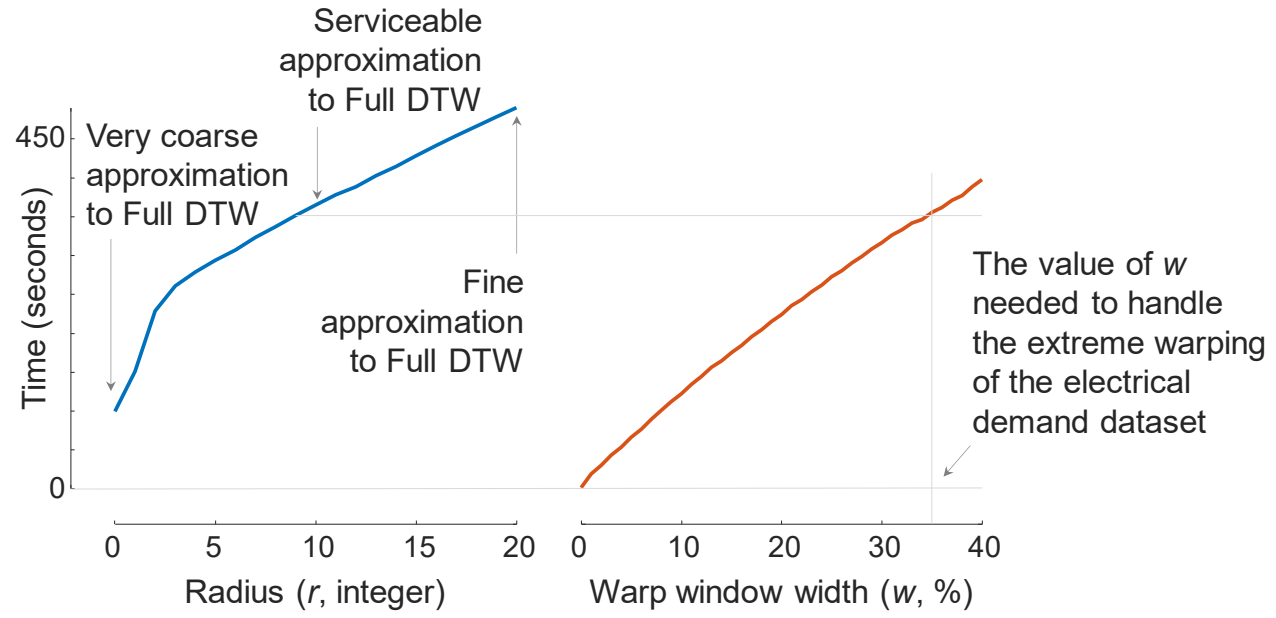

Fig. 4. The experiment shown in Fig. 1 generalized to consider warping window width up to 40%. The time is the cumulative time needed for all pairwise comparisons in a dataset of 1,000 examples (499,500 comparisons).

A recent paper also observed that FastDTW does not achieve expected time optimization in Case C. The authors conclude "*this is because* (our time series) *are usually short, which makes the cost of* (FastDTW) *exceed the cost of calculating DTW directly.*" [23].

Note that as with Case A, we resisted the temptation to do any of the optimizations available only to cDTW when doing multiple comparisons. This is a straightforward head-to-head comparison. Thus, for Case C we find no evidence of the utility of FastDTW.

### 3.4 Case D: Long *N* and Wide *W*

Case D is the case emphasized by the original authors of the FastDTW paper as the best case for their algorithms. However, they did not show any real-world examples of such datasets, and a (admittedly incomplete) survey of the papers that refence FastDTW does not show any examples in the literature [7], [8], [9], [10], [11], [12], [13],[14], [15], [16], [17], [18].

We claim that there are no practical applications of such comparisons. There are simply no problems for which we need to compare time series of this length with a large value for *w*. Of course, it is hard to prove a negative, but consider:

- Modern electrocardiograms can record data at rates of up to 25,000Hz [24]. However, there have been numerous studies that ask, "*what is the minimum sampling rate we need for* (some cardiological problem)?" The answer is typically around 250Hz [25]. This means that to compare two heartbeats, we need to compare about 120 to 200 datapoints. Does it ever make sense to compare longer regions of ECGs? No. To see why, imagine comparing two one-minute long ECGs. Such traces would have about 100 beats, but it is very unlikely that they would have the exact same number. While DTW is forgiving of misalignments, it must explain all the data. It is never meaningful to compare say ninety-eight heartbeats to one-hundred and three heartbeats. Thus, we believe that all uses of DTW for cardiology are in Case A.
- A recent exhaustive empirical study asked a similar question to the above ECG study in the context of gesture recognition. It was discovered that "*recognition rates for N* = 32, (are) *not significantly different than those delivered by higher rates*" [26]. This reflects a complex twenty-five gesture, user-independent study. Perhaps there is some circumstance in which we need a greater *N* for gestures. Perhaps Asians have more nuanced gestures than the tested Europeans. However, this study (and many similar studies) strongly suggest that there is little utility in comparing more than a few hundred data points for human gestures, gait cycles, sport performances etc.
- As noted above, the UCR archive has 128 datasets, many culled from real-world problems. The longest of these is 2,844. However, even for the handful of long time series, the long length is typically just an artifact of how it was recorded. We can downsample most of this time series by a factor of eight or more, and get an accuracy that is statistically significantly the same.

In summary, to the best of our knowledge there is no evidence that it is ever useful to compare time series with lengths exceeding (conservatively) 1,000. Of course, absence of evidence is not evidence of absence. However, it is clear that at a minimum, this is a very rare case.

Nevertheless, for completeness we *do* test this case. We consider a contrived "fall" dataset.

Suppose that a researcher was investigating falls by having actors wearing a motion capture suit fall over in a safe environment. Further imagine she instructs actors to "*Fall over anytime within two seconds of hearing the beep*". Assume she does not crop and clean the data, but simply measures the distance between two-second snippets, which were recorded at 100Hz. Knowing this, we can assume that in this domain $W \approx 100\%$.

Instead of two seconds, let us generalize to $L$ seconds. As shown in Fig. 5 we created a data generator that creates pairs of time series of length $L$ seconds at 100Hz. One time series has an immediate fall, then the actor is near motionless for the rest of the time. For the other time series, the actor is near motionless until just before $L$ seconds are up, then he falls.

It is clear that for cDTW to align the two falls, we must use cDTW$_{100}$.

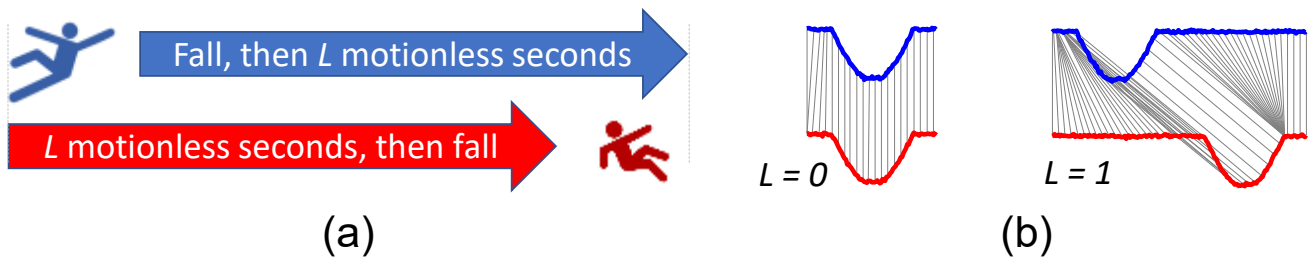

Fig. 5. (a) We model the task of aligning early and late falls in a $L$-second long interval. (b) The cDTW alignment for $L$ = 0 and $L$ = 1, are both examples of Case C, but for large enough values of $L$ we begin to move to Case D.

Note that we do not test to see if FastDTW$_{40}$ actually aligns the two falls, we simply assume it does.

We can now create pairs of time series of increasing values of $L$ and discover at what point FastDTW$_{40}$ becomes as fast as cDTW$_{100}$. Fig. 6 shows the results.

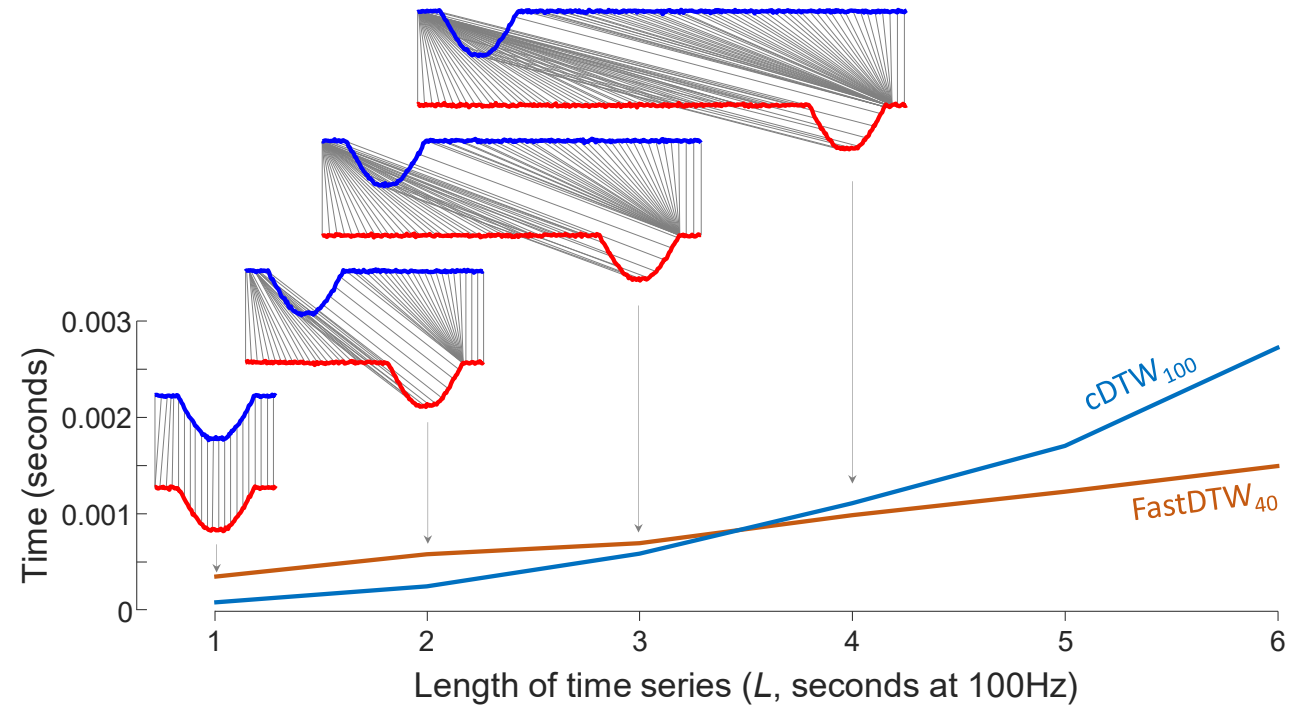

Fig. 6. As we make $L$ longer and longer, we find that when $L$ = 4 ($N$ = 400), FastDTW finally becomes faster than unconstrained cDTW (or cDTW$_{100}$). For each $L$, the time is measured by running each algorithm 1,000 times, and reporting the average.

Thus, we have finally found a circumstance where FastDTW$_{40}$ is faster than cDTW$_{100}$. Note that at this break-even point FastDTW$_{40}$ is an *approximation* to cDTW$_{100}$, so cDTW$_{100}$ is still preferable. However, as $L$ grows well beyond the transition point, each user needs to consider the tradeoff between the time taken vs. utility of approximation. The full answer to that question is beyond the scope of this paper, and in any case depends upon the domain, the analytic task and the cost of an error. For example, if we were comparing some data that we were confident was really in Case D, and we worked out that for the values of $N$, $w$ and $r$, FastDTW$_r$ would be ten times faster, and there was little consequence of using an approximation, we might well decide to use FastDTW$_r$. However, suppose the data in question came from Mars, or from an intrusive, expensive and time-consuming medical biopsy. In these cases, it would be more difficult to justify an approximation, even if it gives you a tenfold speedup.

Another issue is the *magnitude* at which the hypothetical tenfold speedup occurs. There is a real tangible difference between one day and ten days. However, for most practical purposes there is simply no difference between 0.01 seconds and 0.1 seconds. The reader might imagine that *repeated* use of comparisons could allow these short amounts of time to add up to the one/ten days situation: for example, for similarity search or classification. However, for repeated uses of DTW, there are several ideas that can *only* be applied to cDTW, including lower bounding, early abandoning, just-in-time normalization etc. These ideas accelerate cDTW by a further two to five orders of magnitude [3], [4]. For example, for similarity search of a cDTW$_5$ query of length 128, using a 2012 machine, Rakthanmanon *et al.* [3] searched a time series of length one trillion in 1.4 days, however, using a modern machine FastDTW$_{10}$ would take 5.8 years[2]. Likewise, to create the UCR archive [5], Hoang Anh Dau computed cDTW 61,041,100,000,000 times, all on an off-the-shelf desktop, something that would be simply inconceivable with FastDTW.

## 4 When does FastDTW fail to Approximate Well?

In this work we have mostly refrained from measuring the *accuracy* of the FastDTW approximation. Partly this is because we have shown that in almost all cases it is a moot point. In addition, this question opens a pandora's box of what the appropriate measure of quality of approximation is?

Nevertheless, it is instructive to consider one example. We created three time series, and as shown in Table 2, we measured their pairwise distances, using these distance matrices to create the dendrograms shown in Fig. 7.

[2] Averaged over a million comparisons, we found FastDTW$_{10}$ takes 0.1845 milliseconds for $N$ = 128, and $10^{12} \times 0.1845$ milliseconds = 5.8 years.

TABLE 2
The distance matrices for the three time series shown in Fig. 7 under Full DTW and FastDTW$_{20}$

Full DTW

| | A | B | C |
|---|---|---|---|
| A | 0 | 0.020 | 6.822 |
| B | | 0 | 6.848 |
| C | | | 0 |

FastDTW$_{20}$

| | A | B | C |
|---|---|---|---|
| A | 0 | 31.24 | 6.822 |
| B | | 0 | 6.848 |
| C | | | 0 |

The two time series A and B clearly require significant warping. However, as Fig. 7 (c) shows, given unconstrained freedom to warp, they are almost identical, differing only by 0.02. However, FastDTW$_{20}$ finds them to be 31.24 apart. Using the error metric proposed in the original FastDTW paper [6], this is an error of 156,100%.

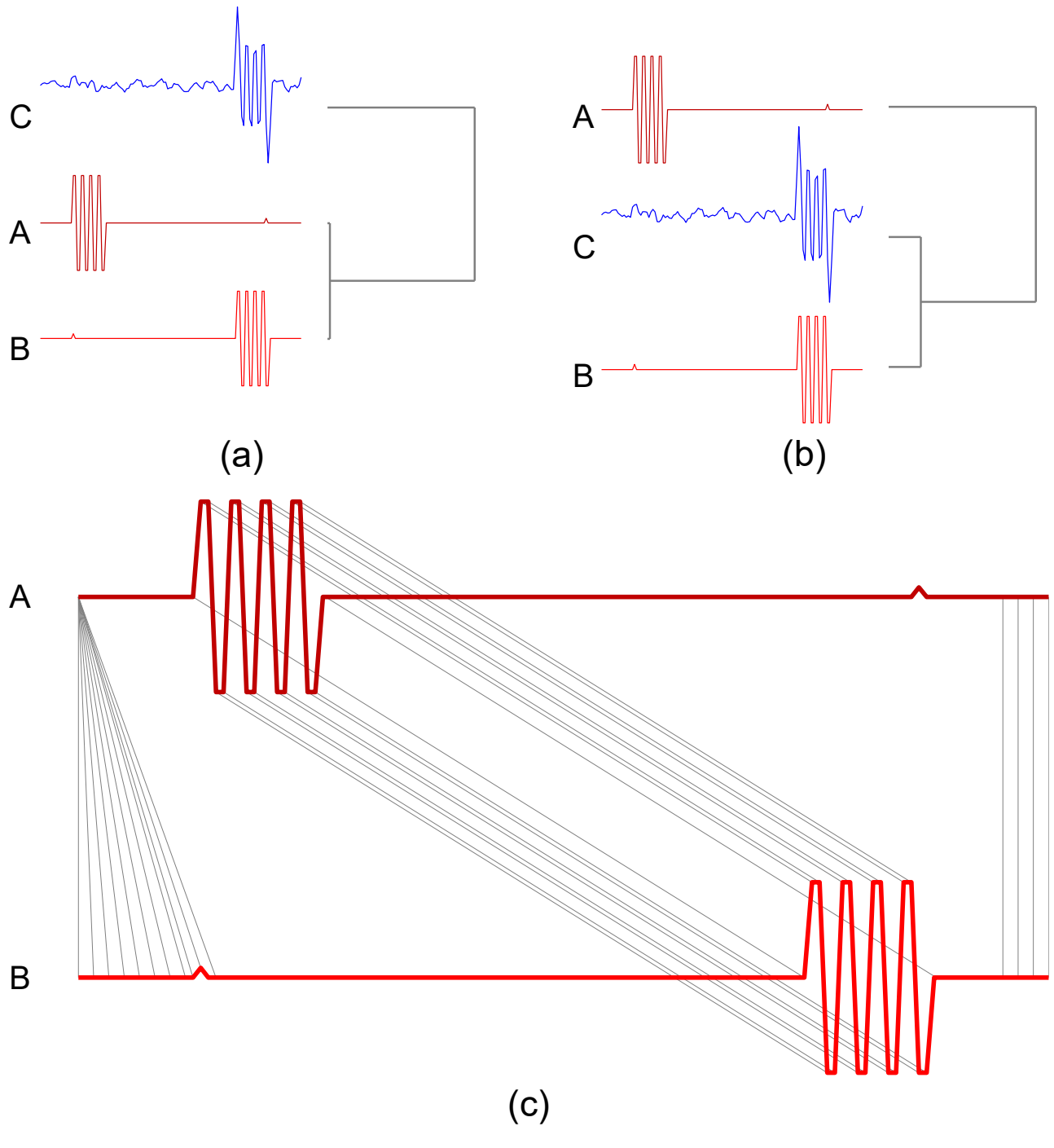

Fig. 7. A clustering of three time series under Full DTW (a) and under FastDTW$_{20}$ (b). (c) The Full DTW alignment between A and B (only selected hatch lines are shown for clarity) shows that they are almost identical if we allow unconstrained warping.

In a sense, this example is unfair to FastDTW$_{20}$. For any distance measure or approximation or upper/lower bound to a distance measure, if you understand how the technique works, you can create synthetic examples that will defeat it. This typically says little or nothing about how likely you are to encounter such adversarial examples in the real world. However, we show this example to make the following point. There appears to be no literature that considers under what conditions FastDTW can fail (and therefore, when to avoid using it). Without such an understanding, practitioners (who are in Case D) may wish to step back and reexamine the trade-offs they are making.

## 5 Summary

We have shown that the vast majority of the researchers that used FastDTW would have been better off simply using cDTW. They would have found simple cDTW to be both faster and to produce exact results.

This paper was not written as part of a game of one-upmanship. It really is the case that there are lost opportunities here, and the community should be aware of this issue. Consider the recent paper [27] which notes "*We employ the FastDTW method by Salvador and Chan…*". The paper shows promising results in gesture recognition, but then ends on the pessimistic question, "*how* (can) *this method can be sped up, desirably up to the point where it reaches real-time capability*". However, for at least the last decade, it was already possible to achieve at least ten thousand times faster real-time performance on their task[3].

## 6 Conclusions

We have shown that a commonly used tool to accelerate time series data analytics does not actually achieve speedup in any realistic setting. We discovered this issue because Salvador and Chan took enormous efforts to make their code available, to clearly explain their approach in their paper, and because they were incredibly responsive to the many questions we asked them. We are extremely appreciative of their assistance.

## Appendix A
## When FastDTW Fails

The result shown in Fig. 7 struck some early readers of this paper as so extraordinary they assumed it was an error on our part. Thus, for completeness, we show how we made this example.

FastDTW assumes that the low dimensionality version of a time series has the same basic shape as the raw data under Piecewise Aggregate Approximation (PAA). This is a reasonable assumption, but as with any dimensionality reduction technique, the pigeonhole principle tells us that there must be examples of objects that are poorly represented in lower dimensionality representation. Suppose that poor approximation for a pair of objects has the property that DTW warps it in the opposite direction to the original data. As Fig. 8 shows, our pair of time series have exactly that property.

[3] Schneider *et al.* [27] conclude by bemoaning the inability to do real time gesture monitoring with FastDTW. They have 36 channels corresponding to different body parts sampled at 30Hz. Let us assume the longest gesture takes two seconds. Can we monitor thirty-six 60-datapoint queries in real time under DTW? Eight years ago, Rakthanmanon *et al.* [3] showed they could monitor a 128-datapoint query at about 6 million Hz. To demonstrate this visually, they produced a video [22] that shows they could monitor a query heartbeat of length 421 at about thirty thousand times faster than real time.

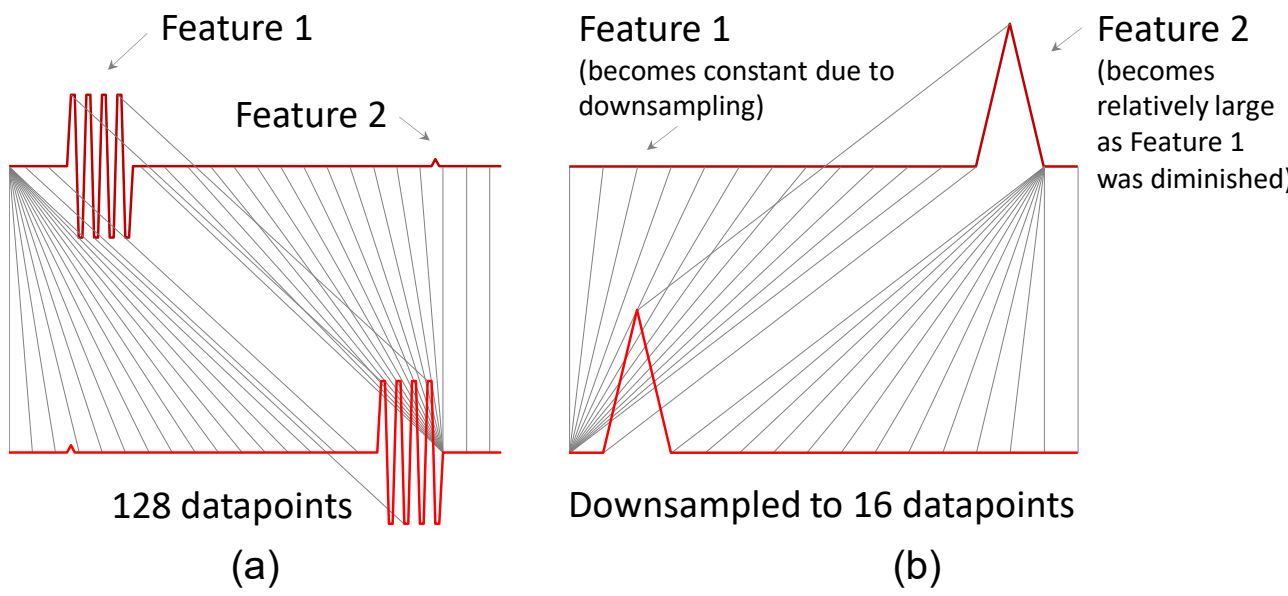

Fig. 8. (a) The two time series shown in Fig. 7 optimally warped by DTW. (b) The eight-to-one PAA downsampled version of the time series depresses the important features and (relatively) magnifies a tiny feature that warps in the opposite direction to the original time series. It is this “wrong way” warping that is passed up to a finer resolution for refinement.

Once the low resolution approximation of FastDTW has committed to warping in the wrong direction, it cannot recover in the higher resolutions, because the parameter *r* excludes reaching the correct warping path.

## APPENDIX B

### INDEPENDENT CONFIRMATION OF OUR CLAIMS

We wrote to several authors that had recently used FastDTW and asked them if they would be willing to rerun their experiments using cDTW. At the time of going to press, we received one reply, from the authors of [27]. Below is the reply, edited for brevity (full text at [28]).

> *I reran the main experiment from our paper with the newest released version of the "fastdtw" package for Python (version 0.3.4) as well as with the implementation you provided me (using radius=30 for both).*
>
> *- Using FastDTW reproduced the same results we originally published (77.38% of gestures correctly classified)*
>
> *- Using your version improved the results of our classifier by about 5% (82.14% of gestures correctly classified)*
>
> *I also compared the runtime of the two DTW implementations during the experiment. I ran the whole experiment twice, which amounts to a total of 2 x 5.851 = 11.702 runs of each DTW implementation for which I compared the runtimes.*
>
> *Result:*
>
> *- your implementation was approx. 24x faster than FastDTW on average (mean: 23.7059, std: 3.587)*
>
> *- in the "slowest" case, your implementation was still approx. 5.8x faster than FastDTW*
>
> *I'd conclude that these tests suggest that your implementation is indeed superior in terms of speed as well as for usage in time-series classification.*

Thus, we have at least one confirmation from a third party that our claims are correct.

## ACKNOWLEDGMENT

We wish to thank Stan Salvador and Philip Chan who corrected several errors in our original understanding of their work and were generous with their time in suggesting experiments and papers to read [21]. Our appreciation does not imply that they endorse this paper.

## REPRODUCIBILITY STATEMENT

We have taken the greatest care to ensure that all experiments in this paper are easily reproducible. To that end, all datasets and code use in this paper are archived in perpetuity at [29]. In the event that we ever discover an issue with this paper that makes us temper its claims slightly, we will discuss it on this page within forty-eight hours. If we ever discover an issue with this paper that significantly affects its claim, we will move to *retract* the paper, but will leave all materials on the website to document our error in perpetuity.

**Renjie Wu** is currently a PhD candidate in Computer Science at the University of California, Riverside. He received his B.S. degree in Computer Science and Technology from Harbin Institute of Technology at Weihai in 2017. His research interests include time series data mining and machine learning.

**Eamonn Keogh** is a professor of Computer Science at the University of California, Riverside. His research interests include time series data mining and computational entomology. He has published thirty-two papers with *DTW* or *Time Warping* in their title.